\documentclass[10pt,twocolumn,letterpaper]{article}
\usepackage{multirow}
\usepackage{subfigure}
\usepackage{booktabs} 
\usepackage[ruled]{algorithm2e}
\usepackage{graphicx}
\usepackage{enumitem}
\usepackage{xcolor}
\usepackage{authblk}
\usepackage{cite}
\usepackage{float}

\def\BibTeX{{\rm B\kern-.05em{\sc i\kern-.025em b}\kern-.08emT\kern-.1667em\lower.7ex\hbox{E}\kern-.125emX}}

\begin{document}

%
\title{A blessing in disguise: Designing Robust Turing Test by Employing Algorithm Unrobustness}

%

\author[1]{Jiaming Zhang}
\author[2]{Jitao Sang}
\author[2]{Kaiyuan Xu}
\author[2]{Shangxi Wu}
\author[1]{Yongli Hu}
\author[1]{Yanfeng Sun}
\author[2]{Jian Yu}
\affil[1]{Beijing University of Technology, Beijing, China}
\affil[2]{Beijing Jiaotong University, Beijing, China}
%
\maketitle

%
\begin{abstract}
Turing test was originally proposed to examine whether machine's behavior is indistinguishable from a human. The most popular and practical Turing test is CAPTCHA, which is to discriminate algorithm from human by offering recognition-alike questions. The recent development of deep learning has significantly advanced the capability of algorithm in solving CAPTCHA questions, forcing CAPTCHA designers to increase question complexity. Instead of designing questions difficult for both algorithm and human, this study attempts to employ the limitations of algorithm to design robust CAPTCHA questions easily solvable to human. Specifically, our data analysis observes that human and algorithm demonstrates different vulnerability to visual distortions: adversarial perturbation is significantly annoying to algorithm yet friendly to human. We are motivated to employ adversarially perturbed images for robust CAPTCHA design in the context of character-based questions. Three modules of multi-target attack, ensemble adversarial training, and image preprocessing differentiable approximation are proposed to address the characteristics of character-based CAPTCHA cracking. Qualitative and quantitative experimental results demonstrate the effectiveness of the proposed solution. We hope this study can lead to the discussions around adversarial attack/defense in CAPTCHA design and alsp inspire the future attempts in employing algorithm limitation for practical usage.
\end{abstract}

%

%
\maketitle

\begin{figure}[t]
  \begin{center}
    \includegraphics[width=\linewidth]{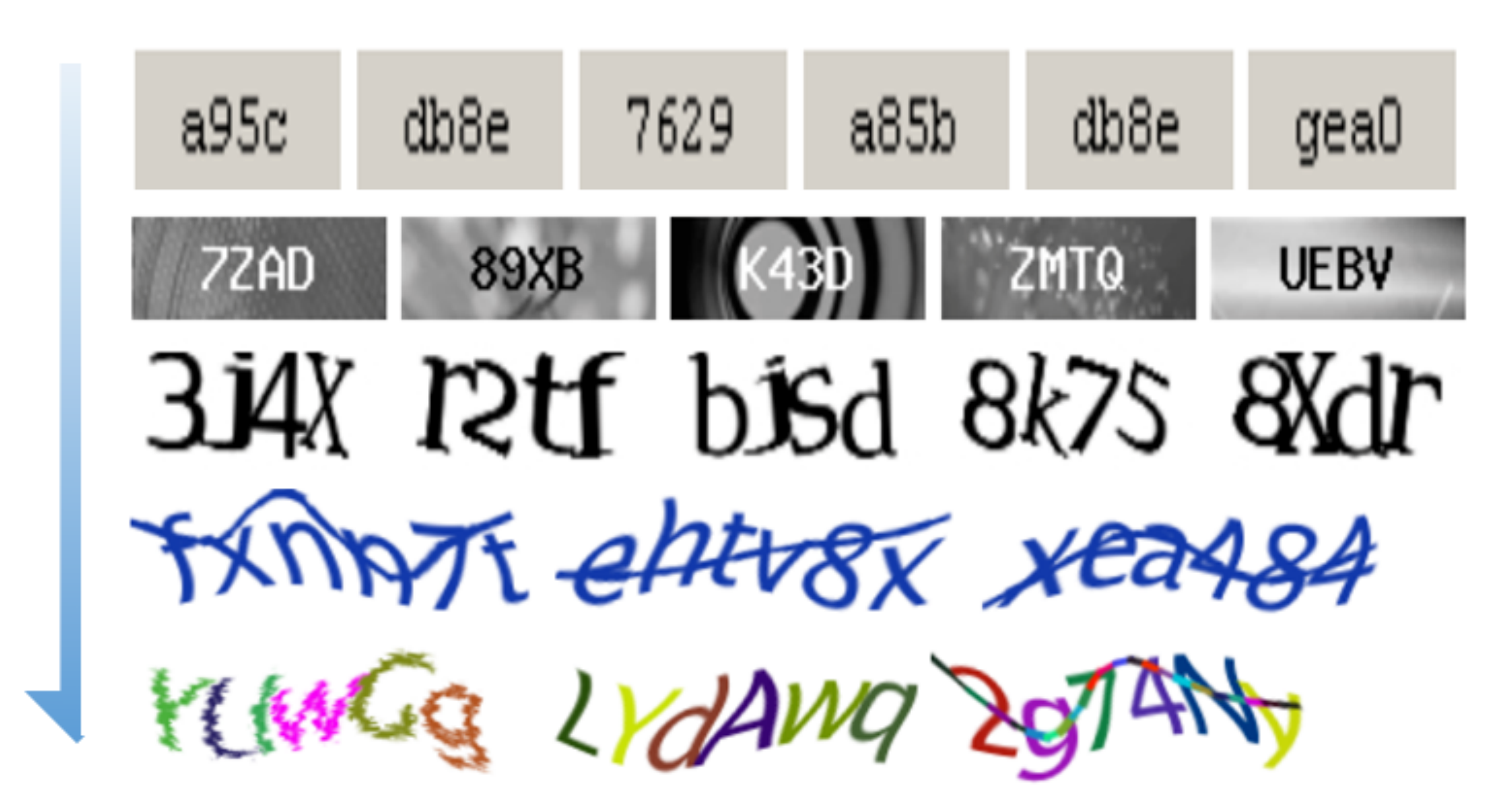}
  \end{center}
  \caption{Increasing content complexity of CAPTCHAs.}
  \label{fig0}
\end{figure}

\section{Introduction}\label{sec1}
Alan Turing first proposed the Turing Test question ``Can machines think like human?''~\cite{machinery1950computing} Turing test was initially designed to examine machine's exhibited intelligent behavior that is indistinguishable from a human, and later developed into a form of reverse Turing test with more practical goal of distinguishing between computer and human. Among reverse Turing tests, CAPTCHA (Completely Automated Public Turing test to tell Computers and Humans Apart) turns out the most well-known one used in anti-spam systems to prevent abuse use of automated programs~\cite{naor1996verification}.

Most early CAPTCHAs, like the reCAPTCHA~\cite{von2008recaptcha} which assists in the digitization of Google books, belong to the traditional character-based scheme involving with only numbers and English characters. With the fast progress of machine learning especially deep learning algorithms, simple character-based CAPTCHAs fail to distinguish between algorithm and human~\cite{chandavale2009algorithm}\cite{sivakorn2016robot}\cite{mori2003recognizing}. CAPTCHA designers were therefore forced to increase the complexity of content to be recognized. As shown in Fig.~\ref{fig0}, while the extremely complex CAPTCHAs reduce the risks to be cracked by algorithms, it also heavily increase the burden of human recognition. It is noteworthy that the effectiveness of simply increasing content complexity is based on the assumption that human has consistently superior recognition capability than algorithm. The last few years have witnessed human-level AI in tasks like image recognition~\cite{he2015delving}, speech processing~\cite{gemmeke2017audio} and even reading comprehension~\cite{Rajpurkar2016SQuAD}. It is easy to imagine that with the further development of algorithms, continuously increasing content complexity will reach such a critical point that algorithm can recognize yet human cannot recognize.

Let's review the initial goal of CAPTCHA: to discriminate human from algorithm by designing tasks unsolvable to algorithms. Therefore, the straightforward solution is to employ the limitations of algorithms to facilitate CAPTCHA design. While algorithms have advanced their performance in many perspectives including visual/vocal recognition accuracy, they remain some notorious limitations with regards to human~\cite{marcus2018deep}. Researchers and practitioners already employed such limitations to design new form of CAPTCHAs, e.g., developing cognitive~\cite{ogiela2018application} and sequentially related~\cite{geman2015visual} questions to challenge algorithm's lack of commonsense knowledge and poor contextual reasoning ability.

\begin{figure*}[t]
\begin{minipage}[b]{0.49\linewidth}
\centering
\includegraphics[width=0.8\textwidth]{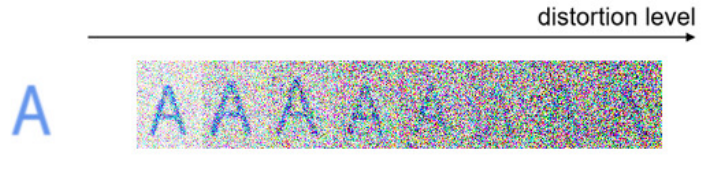}
\centerline{(a)Character distorted by Gaussian white noise}
\end{minipage}
\begin{minipage}[b]{0.49\linewidth}
\centering
\includegraphics[width=0.8\textwidth]{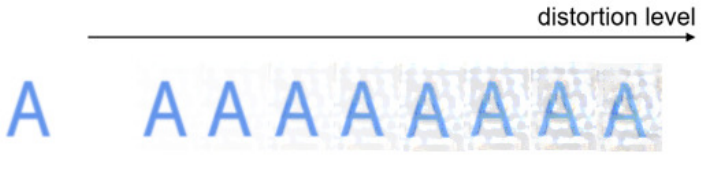}
\centerline{(b)Character distorted by adversarial perturbation}
\end{minipage}
\vspace{1mm}
\begin{minipage}[b]{0.49\linewidth}
\centering
\includegraphics[width=0.89\textwidth]{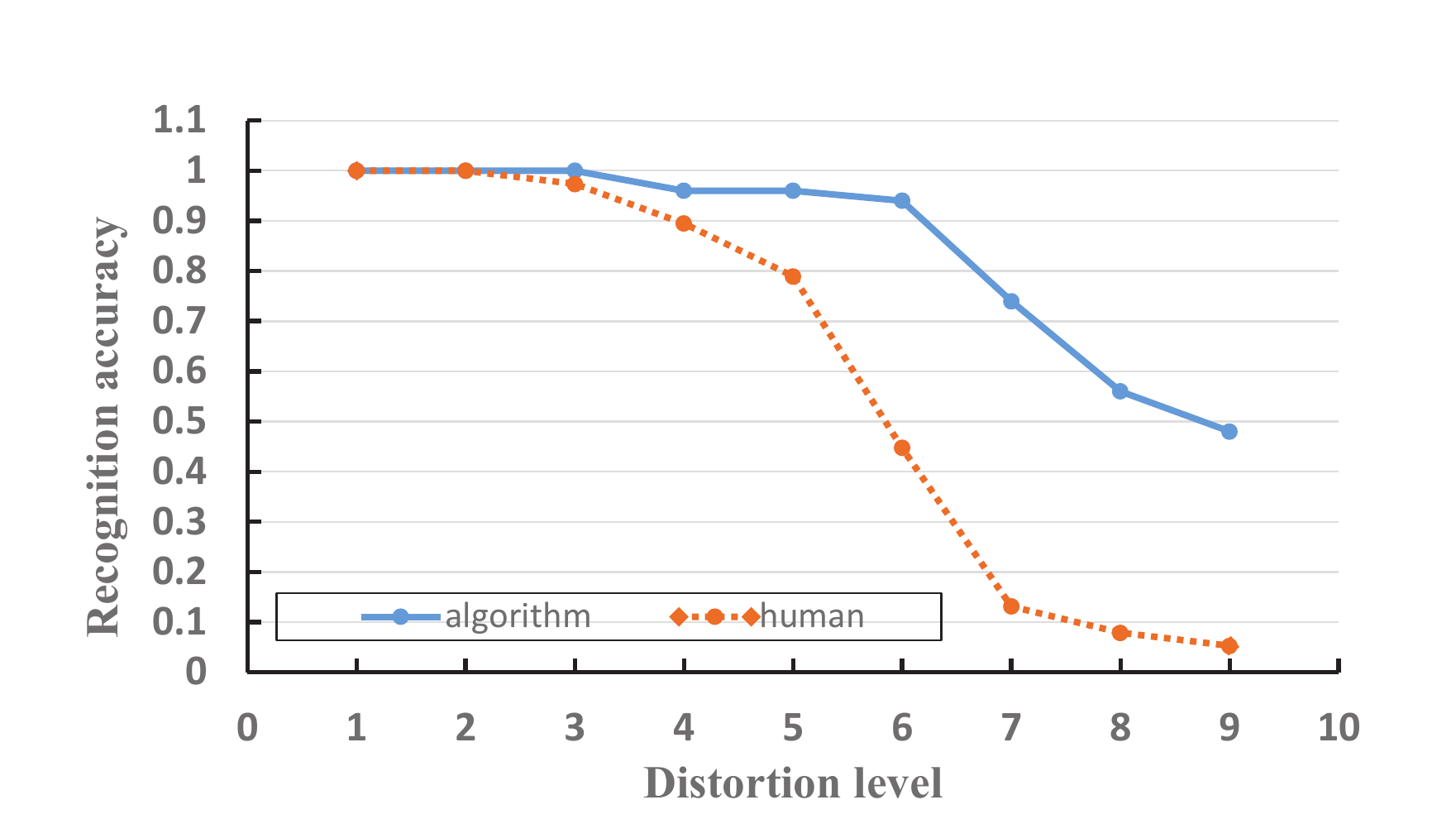}
\centerline{(c)Accuracy on Gaussian distorted characters}
\end{minipage}
\begin{minipage}[b]{0.49\linewidth}
\centering
\includegraphics[width=0.89\textwidth]{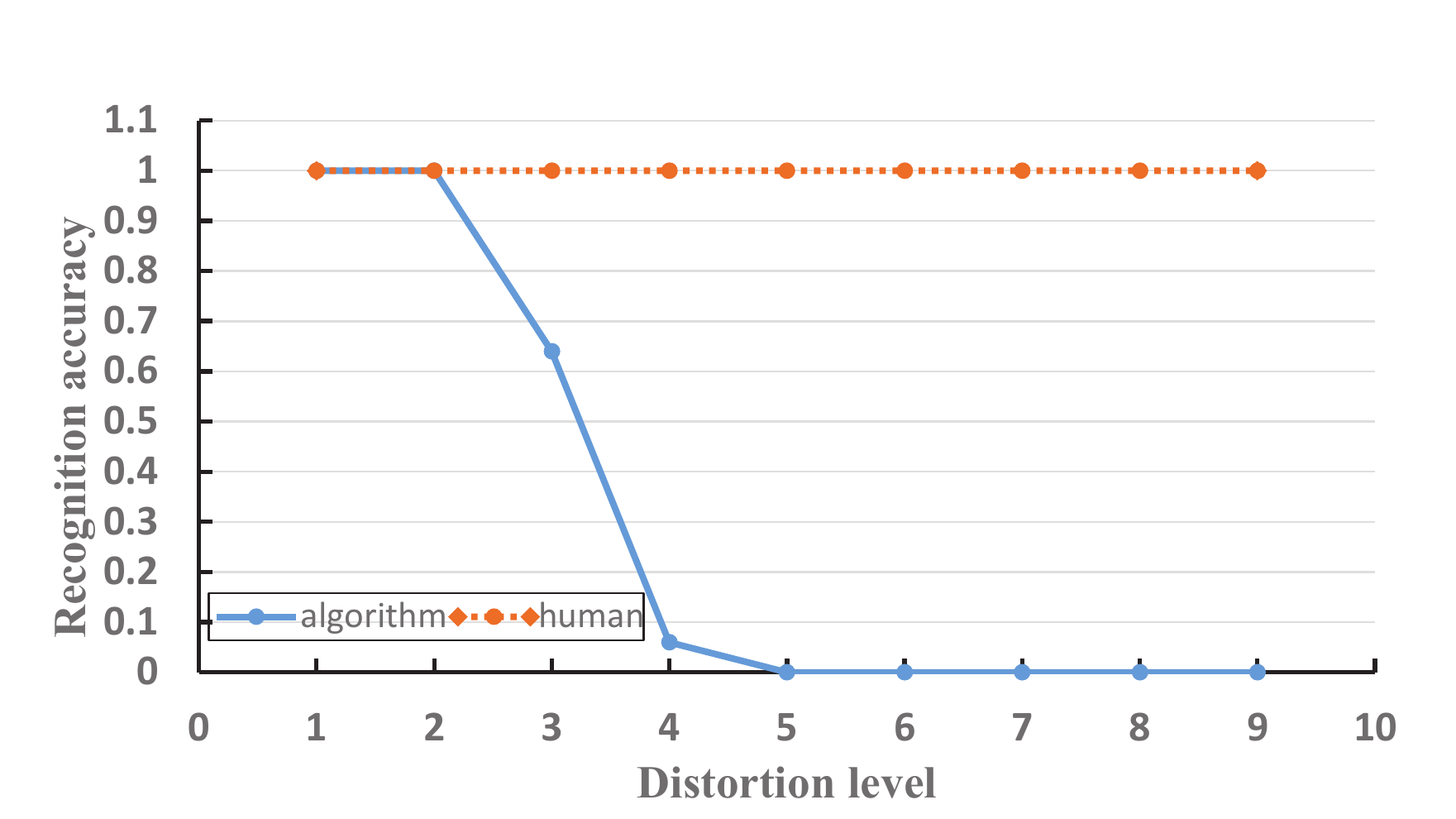}
\centerline{(d)Accuracy on adversarially distorted characters}
\end{minipage}
\caption{Human v.s. algorithm vulnerability analysis results on Gaussian and adversarial distortions}
 \label{fig2}
 \end{figure*}
 
\begin{figure}[t]
\begin{minipage}[b]{0.99\linewidth}
\centering
\includegraphics[width=0.99\textwidth]{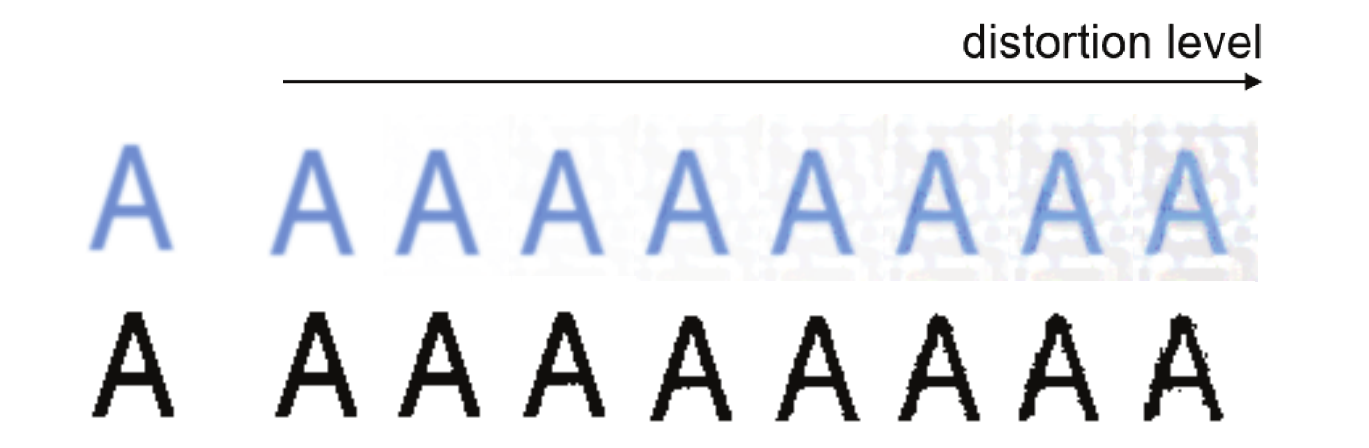}
\centerline{(a)}
\end{minipage}
\begin{minipage}[b]{0.99\linewidth}
\centering
\includegraphics[width=0.99\textwidth]{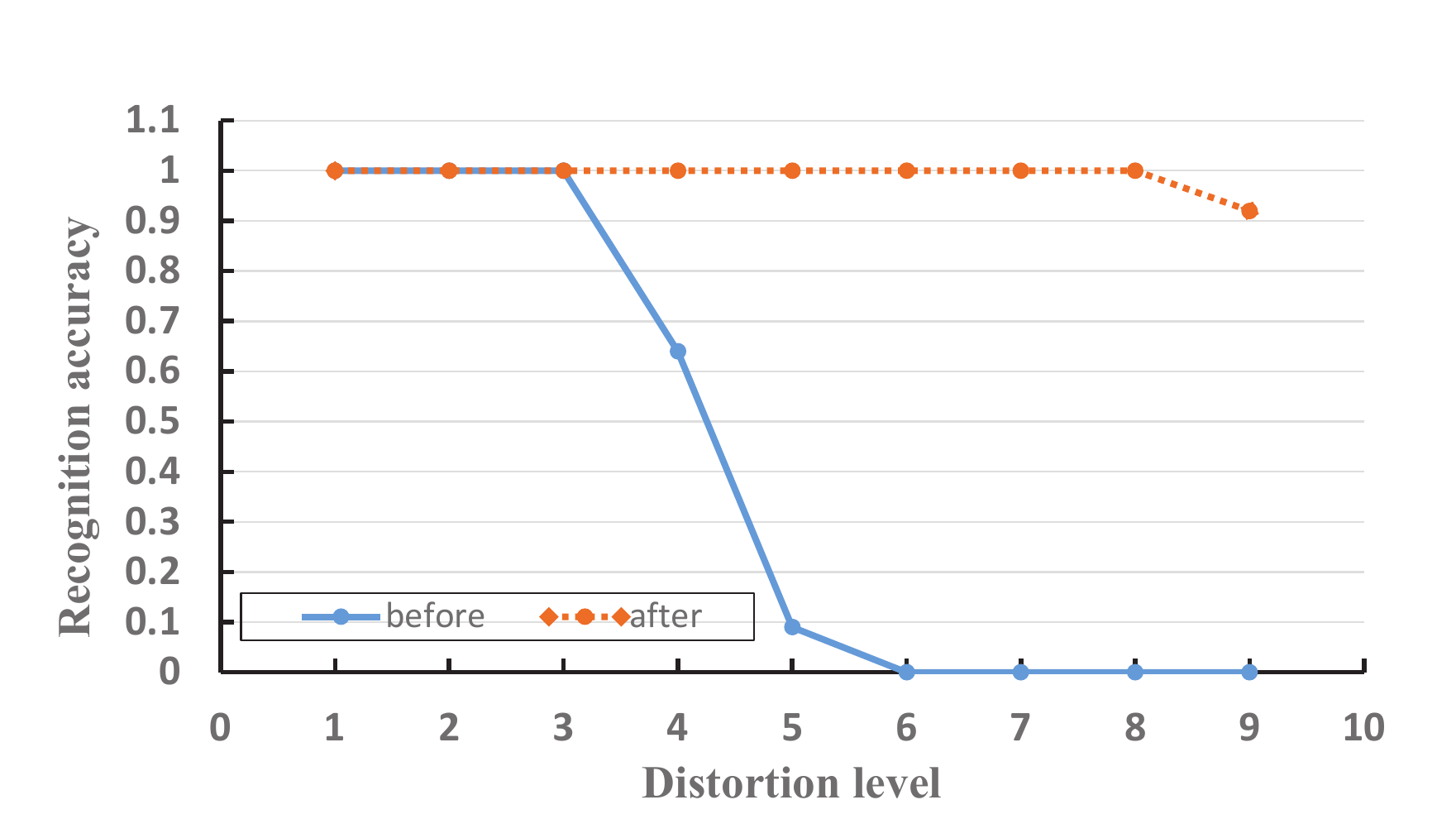}
\centerline{(b)}
\end{minipage}
\caption{The affection of image preprocessing: (a) distorted characters before (top row) and after (bottom row) image binarization; (b)Recognition accuracy on adversarially distorted characters.}
 \label{fig3}
 \end{figure}

Following this spirit, we are interested to explore the possibility of improving the robustness of CAPTCHA towards algorithm cracking without changing the traditional character-based scheme. In other words, is it possible to design character CAPTCHAs only friendly to human instead of simply increasing content complexity? The key lies in finding the algorithm limitation compatible to the scheme of character image. One candidate is the vulnerability to visual distortions. We have conducted data analysis and observed that human and algorithm exhibit different vulnerability to visual distortions (the observations are detailed in Section~\ref{sec2}). This inspires us to exploit those distortions friendly to human but obstructing algorithm to pollute the original character CAPTCHA. Specifically, adversarial perturbation~\cite{szegedy2013intriguing} exactly meets this requirement: adversarial attack~\footnote{~\small{Adversarial attack refers to the process of adding small but specially crafted perturbation to generate adversarial examples misleading algorithm. To avoid confusion with the process of attacking CAPTCHA, in this study, we use ``adversarial attack'' to indicate the generation of adversarially distorted CAPTCHAs and use ``CAPTCHA crack'' to indicate the attempt of passing CAPTCHA with algorithms.}} and CAPTCHA share the common intention that human are imperceptible to but algorithm are significantly affected by the same distortion. The notorious characteristic of adversarial perturbation for visual understanding turns out to be the desired one for CAPTCHA design.

Inspired by this, we employ adversarial perturbation to design robust character-based CAPTCHA in this study. Current state-of-the-art cracking solution views CAPTCHA OCR(Optical Character Recognition) as a sequential recognition problem~\cite{graves2006connectionist}\cite{breuel2013high}. To remove the potential distortions, further image preprocessing operations are typically added before OCR. Correspondingly in this study, we propose to simultaneously attack multiple targets to address the sequential recognition issue (Section~\ref{sec3-1}) and differentiably approximate image preprocessing operations in the adversarial example generation process to cancel out their potential influence (Section~\ref{sec3-3}). Moreover, since we have no knowledge about the detailed algorithm the cracking solution used (e.g., neural network structure), the generated adversarial examples are expected to be resistent to unknown OCR algorithms in the black-box cracking. This study resorts this issue to ensemble adversarial training by generating adversarial examples effective towards multiple algorithms (Section~\ref{sec3-2}). In summary, the contributions of this study are two-fold:
\begin{itemize}
\item We have discovered the different vulnerability between human and algorithm on visual distortions. Based on the observations, adversarial perturbation is employed to improve the robustness of character-based CAPTCHA.
\item Corresponding to the characteristics of typical OCR cracking solutions, we proposed a novel methodology addressing issues including sequential recognition, indifferentiable image preprocessing, and black-box cracking.
\end{itemize}


\section{Data Analysis}\label{sec2}
To justify the feasibility of employing algorithm limitations for CAPTCHA design and motivate our detailed solution, this section conducts data analysis to answer two questions: (1) Whether human and algorithm have different vulnerability to visual distortion? (2) What characteristics to consider when employing distortions to design robust CAPTCHA?

Text-based CAPTCHA is the most widely deployed scheme requiring subjects recognize characters from \emph{0-9} and \emph{A-Z}. Due to its simplicity, character-based CAPTCHA is very effective to examine the robustness towards cracking algorithm as well as friendliness to human. Therefore, this study employs character-based CAPTCHA as the example scheme to conduct data analysis, develop solution and implement experiments. Specifically, during data analysis, we assume that each CAPTCHA question is constituted by single character in a RGB image with unique resolution of $48 \times 64$\emph{px}. The character font is fixed as \emph{DroidSansMono}. The remainder of the section will report the observations regarding human and character recognition performance in different scenarios.

\subsection{Vulnerability Analysis to Visual Distortion}\label{sec2-1}
This subsection designs character recognition competition between human and algorithm to analyze their vulnerability to visual distortions. We employed two types of visual distortions: (1) Gaussian white noise is one usual distortion to generate CAPTCHAs~\cite{carter2017op}. In this study, the added one-time Gaussian white noise follows normal distribution with mean $\tilde{\mu}=0$, variance $\tilde{\sigma}=0.01$ and constant power spectral density. (2) Adversarial perturbation has been recognized as imperceptile to human but significantly confusing algorithm. We employ the widely used $FGSM$~\cite{goodfellow2014explaining} to add adversarial perturbation, where one-time perturbation is constituted with step size of $0.02$. To examine the change of recognition performance with increasing distortion difficulty, we added $8$ levels of distortions onto the original character images accumulatively: each level corresponds to $5$ one-time Gaussian white noises and adversarial perturbations respectively. Examples for derived distorted CAPTCHA images in different levels are illustrated in Fig.~\ref{fig2}(a) and (b).

Regarding the human side, we recruited $77$ master workers from Amazon Mechanical Turk (MTurk). Each subject was asked to recognize $450$ character CAPTCHAs with Gaussian and adversarial distortions in different levels respectively. Regarding the algorithm side, we employed the state-of-the-art OCR (Optimal Character Recognition) algorithm, which is the segmentation-based approach for OCR works by segmenting a text line image into individual character images and then recognizing the characters~\cite{breuel2013high}. The resultant average recognition accuracies for Gaussian and adversarially distorted CAPTCHAs are shown in Fig.~\ref{fig2}(c) and (d). We can see that, for Gaussian distorted CAPTCHAs, human's recognition accuracy consistently declines as the distortion level increases, indicating that Gaussian white noise tends to undermine human's vision. On the contrary, the examined OCR algorithm demonstrates good immunity to Gaussian white noise, possibly due to the noise removal effect by multiple convolutional layers~\cite{liao2018defense}. It is easy to imagine that if we design CAPTCHA by adding Gaussian white noise, as the noise level increases, the resultant CAPTCHAs will critically confuse humans instead of obstructing the cracking OCR algorithms.

For adversarially distorted CAPTCHAs, we observed quite opposite recognition results. Fig.~\ref{fig2}(d) shows that humans are more robust to the adversarial perturbations, while OCR algorithm is highly vulnerable as the adversarial distortion increases. This is not surprising since adversarial perturbation is specially crafted to change the algorithm decision under the condition of not confusing human. This characteristic of adversarial perturbation demonstrates one important limitation of algorithm regards to human ability, which perfectly satisfies the requirement of robust CAPTCHA: algorithm tends to fail, while human remains successful. Therefore, we are motivated to employ adversarial examples to design robust CAPTCHA to distinguish between algorithm and human.


\subsection{Characteristics Affecting Robust CAPTCHA Design}\label{sec2-2}
The previous subsection observes that adversarial perturbation is effective to mislead state-of-the-art OCR algorithm, which shows its potential to be employed to design robust CAPTCHA. However, typical CAPTCHA cracking solution involves beyond OCR, e.g., image preprocessing operations like binarization and Gaussian filtering will be applied to remove distortions before issuing to the OCR module. Fig.~\ref{fig3}(a) illustrates the adversarially distorted CAPTCHA images before and after binarization preprocessing. It is easy to conceive that the effectiveness of adversarial perturbation will be critically affected by image preprocessing operations.

We further quantified this affection by analyzing the OCR performance on the same adversarially distorted CAPTCHA images from previous subsection. The recognition accuracies on the CAPTCHAs before and after binarization preprocessing are plotted and compared in Fig.~\ref{fig3}(b). It is shown that after removing most distortions via image binarization, OCR algorithm demonstrates basically stable performance in recognizing CAPTCHAs with different levels of adversarial perturbation. This tells us that standard adversarial perturbation is insufficient to obstruct the cracking method. It is necessary to design the robust CAPTCHA solution considering the characteristics (like preprocessing operations) of CAPTCHA cracking method.
\section{Methodology}\label{sec3}

\begin{figure*}[ht]
  \centering
  \includegraphics[width=\linewidth]{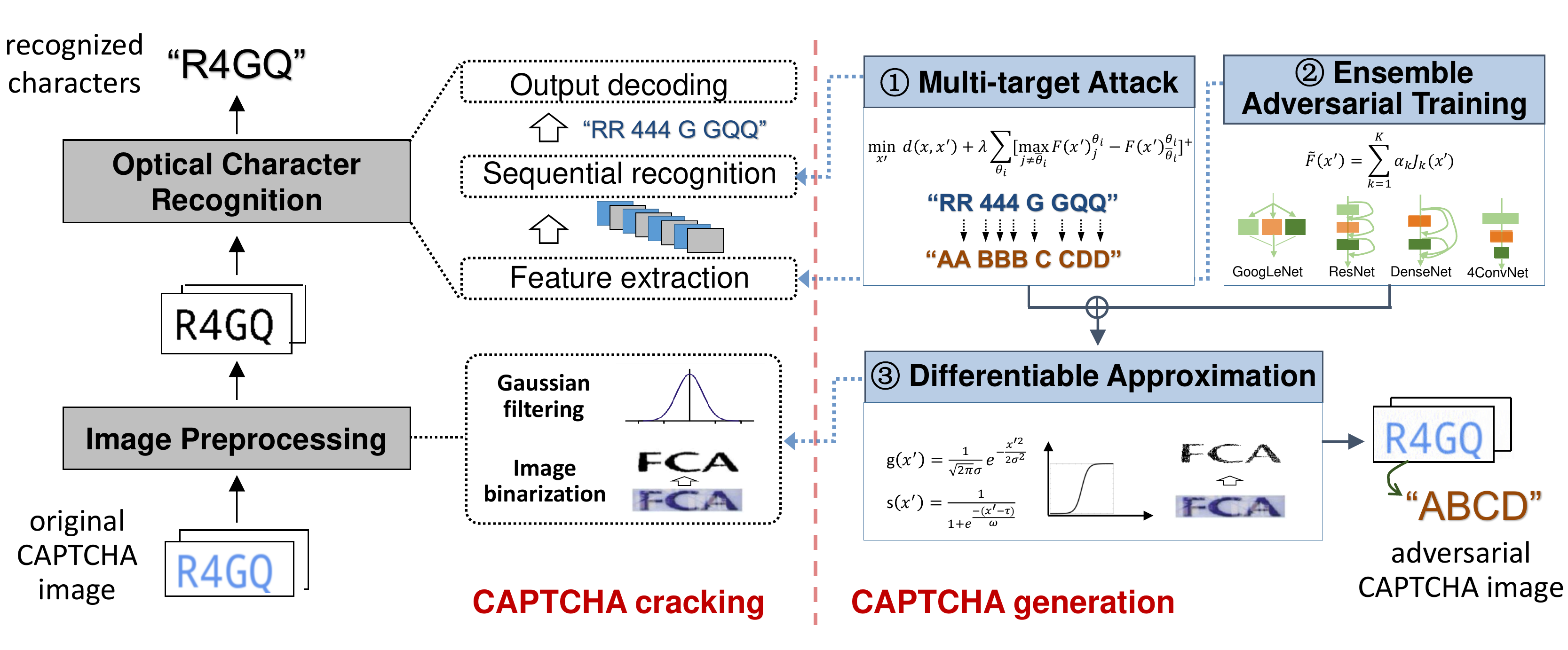}
  \caption{The proposed robust CAPTCHA designing framework.}
  \label{fig:framework}
\end{figure*}

As shown on the left of Fig.~\ref{fig:framework}, typical cracking of character-based CAPTCHA consists of two stages as image preprocessing and OCR. The above data analysis has demonstrated that image preprocessing has the effect of distortion removal, making it not possible to straightforwardly employ adversarial perturbation for robust CAPTCHA design. In addition to the image preprocessing stage, the OCR stage also possesses characteristics obstructing CAPTCHA: (1) sequential recognition, disabling the traditional single character-oriented adversarial perturbation; and (2) black-box crack, making it ineffective to attack one specific OCR model. To address the above characteristics of CAPTCHA cracking, our proposed CAPTCHA generation framework consists of three modules: multi-target attack, ensemble adversarial training, and image preprocessing differentiable approximation. The proposed framework and its relation to CAPTCHA cracking are illustrated on the right of Fig.~\ref{fig:framework}.

%
%
%
%
%
%

\subsection{Multi-target Attack towards Sequential Recognition}\label{sec3-1}
Typical CAPTCHAs usually contain more than one character for recognition, e.g., the example CAPTCHAs contain $4$ characters. Therefore, state-of-the-art CAPTCHA cracking solutions are forced to address a sequential character recognition problem at the OCR stage~\cite{breuel2013high}. Specifically, OCR stage consists of three sub-modules as feature extraction, sequential recognition, and output decoding. Feature extraction is basically realized by a convolutional neural network to encoding the input image as neural feature. Sequential recognition is typically realized by a recurrent neural network to process the issued image neural feature and output multiple tokens including characters (\emph{0-9}, \emph{A-Z}) and blank token $\emptyset$\footnote{~\small{For typical $4$ character-based CAPTCHAs, recurrent neural network usually outputs 12-token sequence to improve tolerance for segmentation and alignment~\cite{breuel2013high}.}} Output decoding serves to transform the sequential tokens into final character recognition results, by merging sequentially duplicated tokens and removing blank $\emptyset$ tokens. For example, the original token sequence ``$aa\emptyset b\emptyset \emptyset ccc\emptyset dd$'' will be transformed to ``$abcd$''.

While CAPTCHA cracking views OCR as a sequential recognition problem, standard adversarial perturbation is designed to attack single target. In this study, we propose to attack multiple targets corresponding to the multiple tokens derived from OCR sequential recognition. The generated adversarial CAPTCHA image is expected to simultaneously misclassify all the character tokens. For specific token sequence $\mathbf{t}$, all the characters appearing in $\mathbf{t}$ constitute the original set $\Theta$, while the remaining characters from (\emph{0-9}, \emph{A-Z}) constitute the adversary set $\bar{\Theta}$. Denoting the raw image as $\mathbf{x}$ and the corresponding adversary image as $\mathbf{x}'$, the multi-target attack is formulated as the following optimization problem:
\begin{equation}\label{eq1}
  \mathop {\min }\limits_{\mathbf{x}'} \quad d(\mathbf{x},\mathbf{x}') + \lambda \cdot \sum_{\theta_i\in{\Theta}}[\max\limits_{j\neq \bar{\theta}_i} F(\mathbf{x}')_{j}^{\theta_i} - F(\mathbf{x}')_{\bar{\theta}_i}^{\theta_i}]^+
\end{equation}
where $d(\cdot,\cdot)$ is distance function to minimize the modification from $\mathbf{x}$ to $\mathbf{x}'$\footnote{~\small{Alternative choices for the distance function are allowed. In our experiment, we use $L_{2}$ distance.}}, $\lambda$ is the weight parameter balancing between the image modification and the misclassification confidence in the second term. Within the second term, $\theta_i$ is the character appearing in the original set $\Theta$, $\bar{\theta}_i$ is its one-to-one mapping character in the adversary set $\bar{\Theta}$, $F(\mathbf{x}')^{\theta_i}$ denotes the output of the second-to-last layer, the logits, corresponding to token $\theta_i$ after sequential recognition, $F(\mathbf{x}')_{j}^{\theta_i}$ denotes its $j$ dimension, and $[f]^+$ is the positive part function denoting $max(f,0)$. Note that the one-to-one mapping from $\theta_i$ to $\bar{\theta}_i$ can be either random or fixed. Random one-to-one mapping leads to targeted adversarial attack, and fixed mapping leads to non-targeted adversarial attack\footnote{~\small{The reported experimental results in Section~\ref{sec4} are based on random one-to-one mapping.}}.

When the original set $\Theta$ contains only one character, the multi-target attack reduces to single-target attack as the standard adversarial perturbation. In fact, according to the mechanism of output decoding in CAPTCHA cracking, we only need to misclassify any one of the character tokens to invalid the final recognition result. The above equation in Eqn.~\ref{eq1} provides a general case of attacking flexible numbers of character tokens. In practice, the number of attacked characters is one importance parameter to control the model performance. More attacked characters guarantee higher success rate to resist crack, yet leading to more derived distortions and human recognition burden. The quantitative influence of attacked character number on the image distortion level and algorithm recognition rate is discussed in Section~\ref{sec4-3}.


\subsection{Ensemble Adversarial Training towards Black-box Crack}\label{sec3-2}
As mentioned in Section~\ref{sec1}, CAPTCHA cracking may employ multiple OCR algorithms for character recognition. At the stage of designing CAPTCHA, it is impractical to target towards one specific OCR algorithm, which require to design adversarial CAPTCHA images that are effective to as many OCR algorithms as possible. Fortunately, it is recognized that adversarial perturbation is transferable between models: if an adversarial image remains effective for multiple models, it is more likely to transfer to other models as well~\cite{papernot2016transferability}. Inspired by this, in order to improve the resistance to unknown cracking models, we propose to generate adversarial images simultaneously misleading multiple models. 

Specifically, given $K$ white-box OCR models with their corresponding the output of the second-to-last layer as $J_{1},...,J_{K}$, we re-formulate the objective function in Eqn.~\ref{eq1} by replacing $F(\mathbf{x}')$ with $\tilde{F}(\mathbf{x}')$ defined as follows:
\begin{equation}\label{eq2}
    \tilde{F}(\mathbf{x}') = \sum^{K}_{k=1} \alpha_{k} J_{k}(\mathbf{x}')
\end{equation}
where $\alpha_{k}$ is the ensemble weight with $\sum^{K}_{k=1}\alpha_{k} = 1$. In most cases, $\alpha_{k} = 1/K$ except that one model is more important than others. Among the three sub-modules of OCR stage, feature extraction has the most model choices (e.g., various CNN structures as GoogLeNet~\cite{szegedy2015going}, ResNet~\cite{he2016deep}) which can be easily implemented into different CAPTCHA cracking solutions. Therefore, this study addresses the black-box cracking issue by attacking multiple feature extraction models. Specifically, the training data and basic structure of $J_{i}(\mathbf{x}')$ and $F(\mathbf{x}')$ are identical except for the different CNN structures in the feature extraction sub-module. On the number of CNN structures, the larger the value of $K$, the stronger the generalization capability of the derived adversarial CAPTCHA images. However, an excessive $K$ value will lead to high computational complexity and trivial weight $\alpha_{k}$ to underemphasize single model. According to previous studies on ensemble adversarial attack~\cite{liu2017delving}, $3\sim5$ models achieve a good balance between transferability and practicality. In this study, we select $K=4$ and evenly set $\alpha_{k} = 1/4$. The performance of employing ensemble adversarial training to resist different OCRs is reported in Section~\ref{sec4-4}.


\begin{table*}[t]
\centering
\caption{The recognition of different complexity levels of CAPTCHAs in the different settings. The results of algorithms are obtained after Gaussian filtering and image binarization.}\label{tab:1}
\begin{tabular}{cccccc} 
\toprule[1pt]
\multicolumn{2}{c}{ }& \emph{Raw}& \emph{rCAPTCHA$\_$parallel}& \emph{rCAPTCHA$\_$w/o preprocessing}& \emph{rCAPTCHA}\\
\midrule[1pt]
\multirow{2}*{Easy}& algorithm& 100.0\%& 95.6\%& 68.4\%& \textbf{0.0\%}\\
\cline{2-6}
 &human &\emph{99.0}\%& \emph{94.0}\%& \emph{94.0}\%& \emph{94.0}\%\\
\midrule[1pt]
\multirow{2}*{Medium}& algorithm& 91.0\%& 88.0\%& 58.0\%&  \textbf{0.0\%}\\
\cline{2-6}
 &human & \emph{73.0}\%& \emph{51.0}\%& \emph{67.0}\%& \emph{65.0}\%\\
\midrule[1pt]
\multirow{2}*{Hard}& algorithm& 81.0\%& 83.0\%& 45.0\%& \textbf{4.0\%}\\
\cline{2-6}
 &human & \emph{56.0}\%& \emph{36.0}\%& \emph{51.0}\%& \emph{49.0}\%\\
\bottomrule[1pt]
\end{tabular}
\end{table*}

\subsection{Differentiable Approximation towards Image Preprocessing}\label{sec3-3}
The data observations in Section~\ref{sec2-2} demonstrate the distortion removal consequences from binarization operation, requiring us to consider the affection of image preprocessing in adversarial image generation. To address this, we regard image preprocessing operation as part of the entire end-to-end solution so that we can generate corresponding adversarial images effective to mislead the whole cracking solution.

According to the usability to be incorporated into the end-to-end solution, image preprocessing operations can be roughly divided into two categories as either differentiable or non-differentiable. For each category, we select one representative operation to address in this study, i.e., Gaussian filtering and image binarization. Regarding the differentiable Gaussian filtering operation, $g(\mathbf{x}')=\frac{1}{\sqrt{2\pi} \sigma}e^{-\frac{\mathbf{x}^{'2}}{2\sigma^{2}}}$, we can readily incorporate it into the OCR model (Eqn.~\ref{eq1}, Eqn.~\ref{eq2}) by replacing the input image $\mathbf{x}'$ with the preprocessing image $g(\mathbf{x}')$. Both forward and backward propagation are conducted on the replaced function ${F}(g(\mathbf{x}'))$, leading to the generated adversarial images expected to eliminate the affection from Gaussian filtering.

Regarding the non-differentiable image binarization , we cannot straightforwardly incorporate it into the objective function. Instead, we find a differentiable approximation $s(\mathbf{x}')$ to image binarization and incorporate the approximated function into the end-to-end solution. In this study, $s(\mathbf{x}')$ is defined as follows:
\begin{equation}\label{eq3}
    s(\mathbf{x}') = \frac{1}{1+e^{-\frac{\mathbf{x}'-\tau}{\omega}}}
\end{equation}
where $\tau$ denotes the threshold of image binarization, $\omega$ denotes the degree of lateral expansion of the curve. Note that to guarantee that the generated adversarial images are resistent to image binarization, we only employ the approximated $s(\mathbf{x}')$ at the backward propagation stage to update the generated image, while the forward propagation still use the actual $\mathbf{x}'$ to calculate $\nabla_{x}F(x)$. 

To simultaneously resist to the affections from Gaussian filtering and image binarization, we concatenate $s(\cdot)$ and $g(\cdot)$ in the final objective function. Therefore, the overall optimization problem incorporating the three proposed modules is as follows:
\begin{equation}\label{eq4}
  \mathop {\min }\limits_{\mathbf{x}'} \quad d(\mathbf{x},\mathbf{x}') + \lambda \cdot \sum_{\theta_i\in{\Theta}}[\max\limits_{j\neq \bar{\theta}_i} \tilde{F}(\phi(\mathbf{x}'))_{j}^{\theta_i} - \tilde{F}(\phi(\mathbf{x}')))_{\bar{\theta}_i}^{\theta_i}]^+
\end{equation}
where $\tilde{F}(\cdot)$ denotes the ensemble of multiple OCR models defined in Eqn.~\ref{eq2}, and $\phi(\mathbf{x}')=s(g(\mathbf{x}'))$ denotes the approximated image preprocessing operations defined in Eqn.~\ref{eq3}.


\section{Experiments}\label{sec4}

We examined CAPTCHA images with $4$ characters for experiments. The CAPTCHAs are RGB images with resolution of $192\times 64px$. Regarding the cracking method, we considered image binarization and Gaussian filtering (kernel size: $3\times 3, \sigma=0.8$) at the image preprocessing stage. The OCR stage is instantiated with CNN structures for feature extraction and LSTM$+$softmax for sequential recognition. Regarding our proposed CAPTCHA generation method, image binarization is approximated with $\tau=0.8, \omega=0.05$, and $4$ CNN structures are employed for ensemble adversarial training. All experiments are conducted on Nvidia GTX 1080Ti GPU with 11G memory.


\begin{figure}[t]
  \begin{center}
    \includegraphics[width=\linewidth]{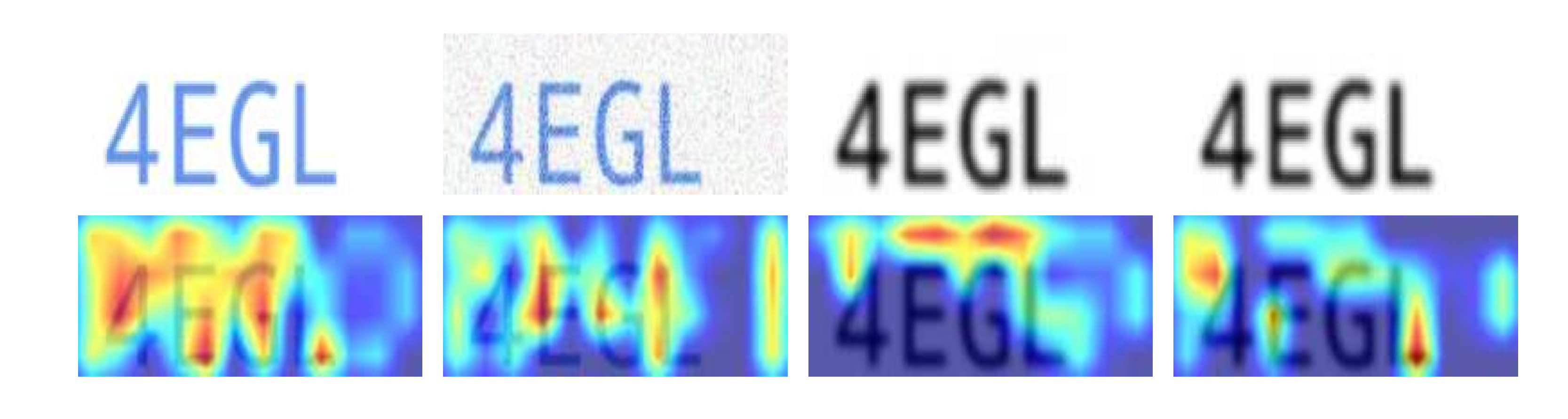}
  \end{center}
  \caption{Example images (top row) and theri attention maps (bottom row). From left to right, we show the original image, the image with Gaussian white noise, the adversarial generated by our method and the adversarial image generated by our method but without considering image preprocessing.}
  \label{fig:attention}
\end{figure}

\subsection{Qualitative Attention Analysis}\label{sec4-1}

Visual attention has been widely used to explain which region of image contributes much to the model decision~\cite{zeiler2014visualizing}. In this study, we extracted the attention map using Grad-CAM~\cite{selvaraju2017grad} to understand the change of recognition performance under different visual distortions.

The first and second columns of Fig.~\ref{fig:attention} visualize the attention map of the raw image and the image with Gaussian white noise. It can be found that Gaussian white noise brings trivial attention change from the original image. Both attention maps keep to the region where characters exist. This well explains the data observations in Section~\ref{sec2-1} that algorithm is generally robust to Gaussian white noise. We also visualized the attention map of the CAPTCHA images generated from our proposed method on the third column of Fig.~\ref{fig:attention}. It is shown that the attention maps deviate much from the original image and focus on unrelated regions where there exist no characters. This justifies our motivation to employ adversarial perturbation to mislead the algorithm prediction result, and demonstrates the effectiveness of our proposed CAPTCHA design method.

To further validate the necessity of considering image preprocessing in robust CAPTCHA design, the attention maps for the images generated from our method but without considering image preprocessing are shown on the fourth column of Fig.~\ref{fig:attention} for comparison. It is easy to conceive that without considering image preprocessing, the generated images fail to deviate the attention from the character regions. This is consistent with the fact that image preprocessing has the effect of weakening or eliminating adversarial perturbation.

\subsection{Quantitative Performance Comparison}\label{sec4-2}

To compare the performance of the proposed robust CAPTCHA (rCAPTCHA) designing method, we report the recognition accuracies of state-of-the-art cracking solution under the following settings: 

\begin{itemize}
\item \emph{Raw}: the original CAPTCHA images without adding adversarial perturbations;
\item \emph{rCAPTCHA$\_$parallel}: the proposed solution to generated adversarial images, expect that the sequential recognition sequential recognition sub-module of OCR is replaced by $4$ parallel recognition networks (each realized by one fully-connected layer) to address one character's recognition;
\item \emph{rCAPTCHA$\_$w/o preprocessing}: the proposed solution to generated adversarial images, but without considering the image preprocessing stages;
\item \emph{rCAPTCHA}: the proposed solution to generated adversarial images, considering both sequential recognition and the image preprocessing operations.
\end{itemize}

\begin{figure}[t]
  \begin{center}
    \includegraphics[width=\linewidth]{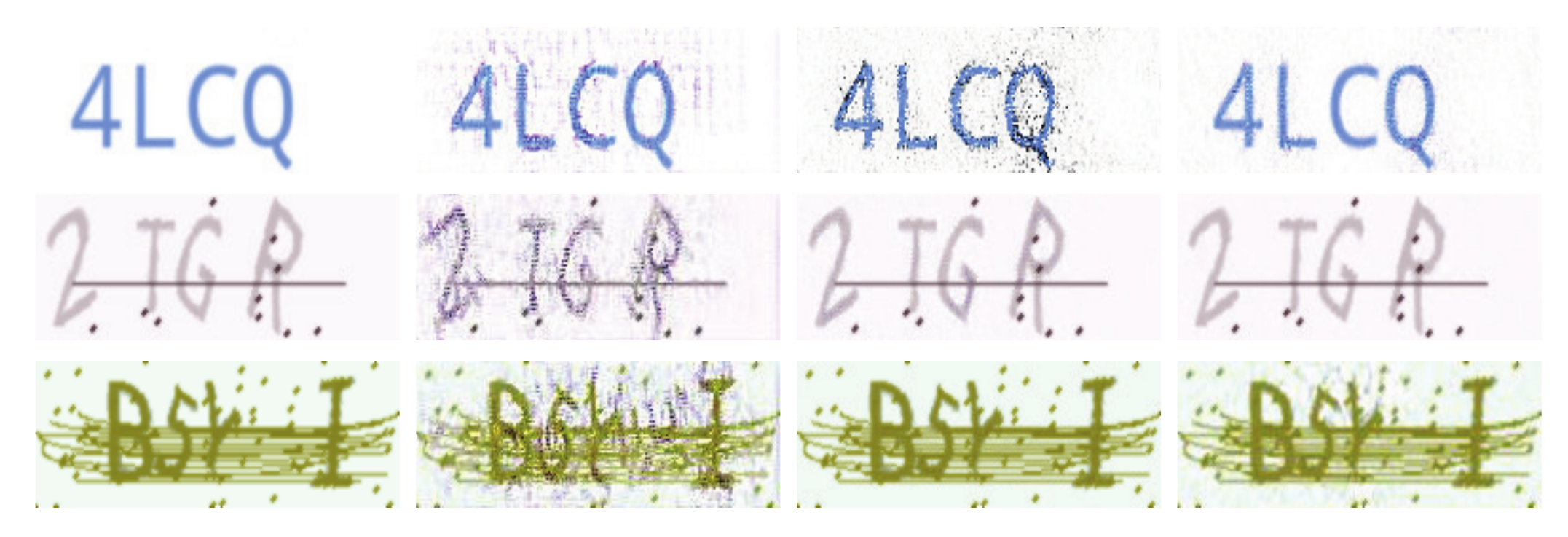}
  \end{center}
  \caption{Example CAPTCHAs with different complexity levels (from top to bottom: {\emph{easy, medium, hard}}). Each row from left to right shows the different settings of Raw, rCAPTCHA$\_$parallel, rCAPTCHA$\_$w/o preprocessing and rCAPTCHA.}
  \label{fig:experiment}
\end{figure}

The state-of-the-art cracking solution is trained over $20,000$ CAPTCHA images with batch size $128$. To examine the application scope of the proposed CAPTCHA generation methods, we conducted experiments on the CAPTCHAs with three levels of complexities: \emph{easy, medium, hard}. Fig.~\ref{fig:experiment} shows examples of different complexity levels of CAPTCHAs in the above four settings. For each of the settings, we selected/generated $500$ CAPTCHA images for testing, and summarize the derived average recognition accuracy in Table~\ref{tab:1}. Experimental observations include: (1) By adding adversarial perturbations, the right $3$ columns consistently obtain lower accuracies than the first column, showing the usability of employing adversarial perturbations in resisting CAPTCHA cracking. (2) Without considering the sequential recognition or image preprocessing characteristics, the resisting effect of \emph{rCAPTCHA$\_$parallel} and \emph{rCAPTCHA$\_$w/o preprocessing} is not as obvious as that of \emph{rCAPCHA}. This validates the necessity of multi-target attack and differentiable approximation modules. (3) Regarding CAPTCHAs with different complexities, we observed consistent phenomenon among the four settings, demonstrating the wide application scope of the proposed CAPTCHA generation method.

The notable decrease in algorithm recognition accuracies shows the effectiveness of employing adversarial perturbation to mislead cracking solution. To facilitate the correlation understanding between misleading cracking solution and friendlity to human recognition, we also provide the human recognition accuracy for each experimental setting in Table~\ref{tab:1}. Similar to the data analysis, we have recruited $164$ workers from MTurk to recognize $4$ character-based CAPTCHA images. The reported accuracies are averaged over $1,200$ CAPTCHAs. By comparing different rows, it is shown that the increasing content complexity brings slight decrease of algorithm recognition accuracy but causes huge trouble to human recognition. Among different setting columns, while the algorithm recognition accuracy fluctuates a lot, the human recognition performance basically remains stable, validating the different distortion vulnerability between human and algorithm. In summary, regarding CAPTCHA images with different complexities, the proposed CAPTCHA generation method succeeds to invalid the cracking algorithm without increasing human recognition burden. 

\begin{figure}[t]
\begin{minipage}[b]{0.99\linewidth}
\centering
\includegraphics[width=1.0\textwidth]{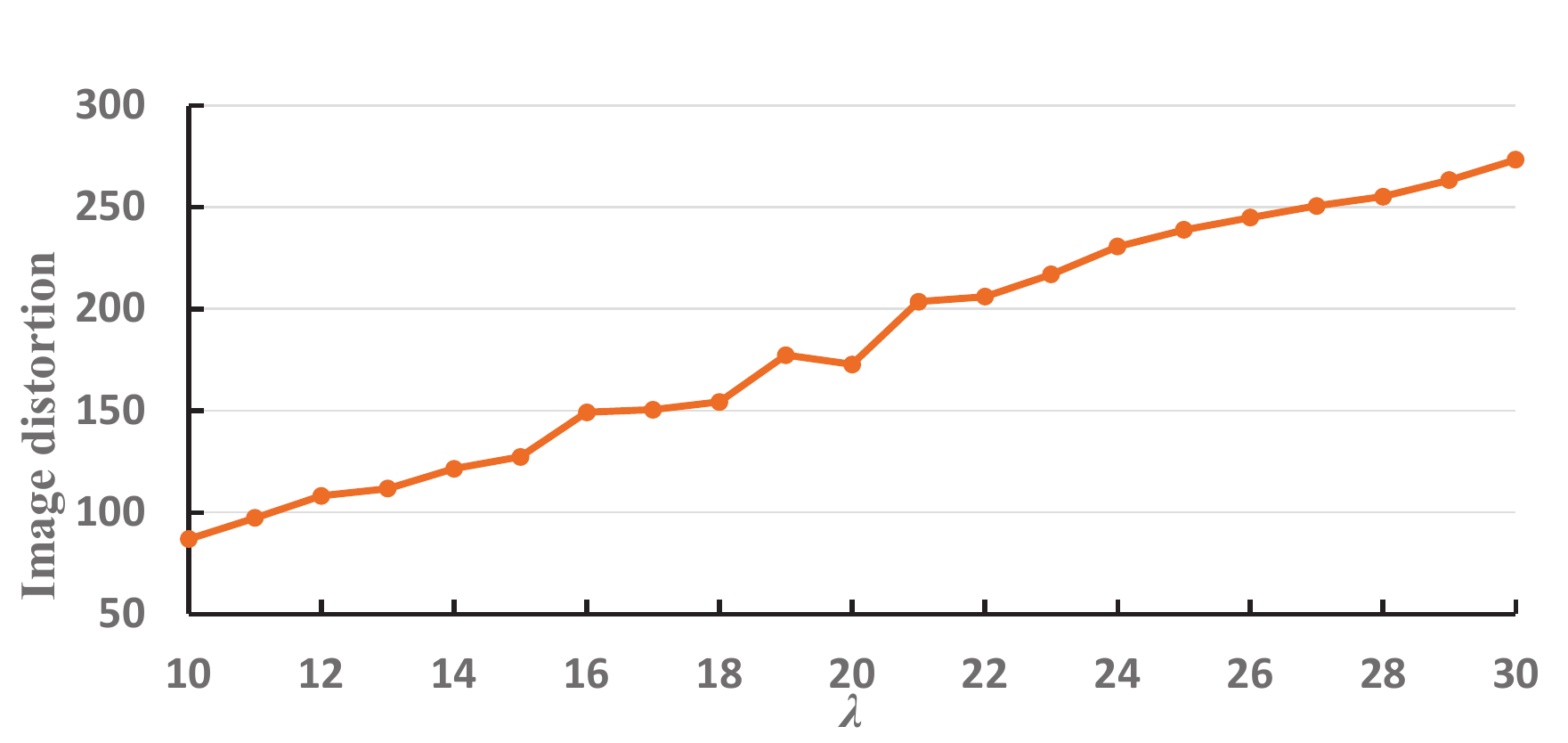}
\centerline{(a) Image distortion}
\end{minipage}
\begin{minipage}[b]{0.99\linewidth}
\centering
\includegraphics[width=1.0\textwidth]{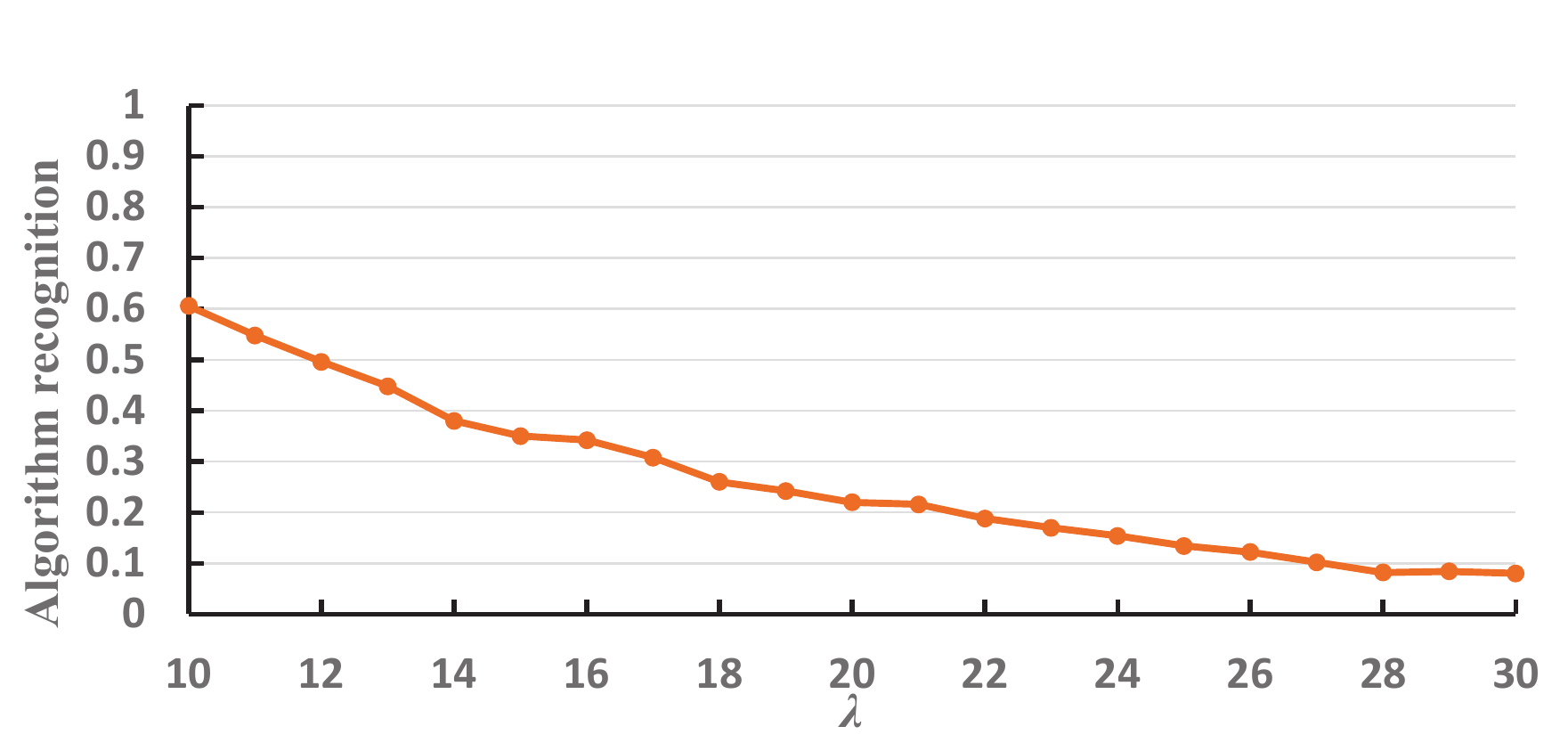}
\centerline{(b) Algorithm recognition accuracy}
\end{minipage}
\caption{The influence of $\lambda$ on derived image distortion and cracking recognition accuracy.}
\label{fig:lambda}
\end{figure}

\begin{figure}[t]
  \begin{center}
    \includegraphics[width=\linewidth]{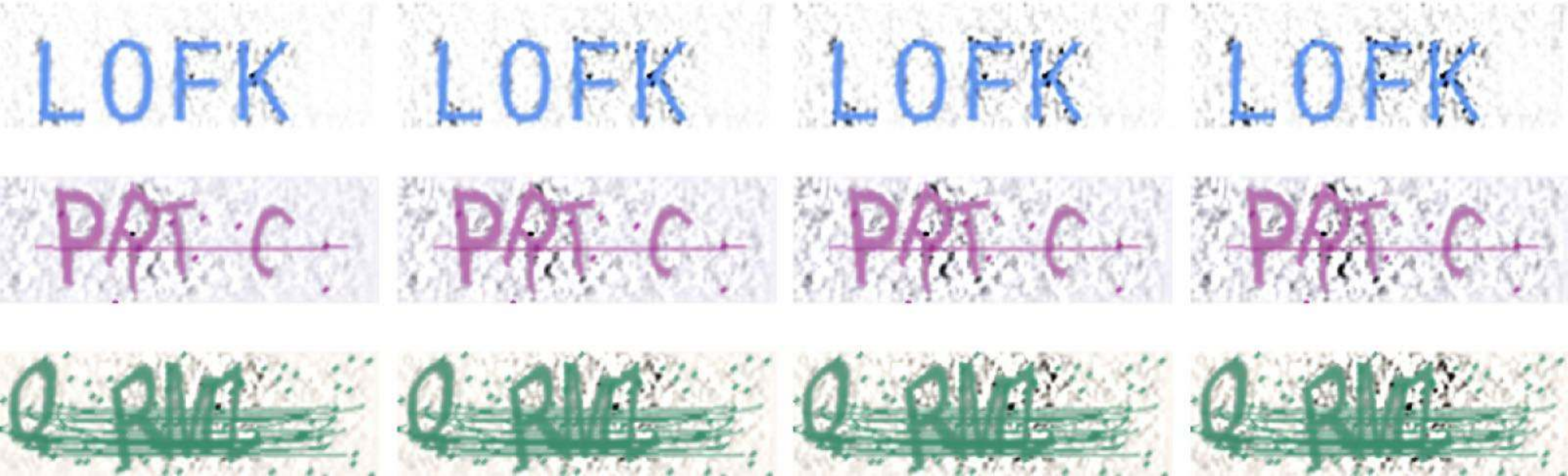}
  \end{center}
  \caption{Example CAPTCHAs with different image distortions: from left to right shows images with distortions of $100$, $200$, $300$ and $400$.}
  \label{fig:distortion}
\end{figure}

\begin{figure}[t]
\begin{minipage}[b]{0.99\linewidth}
\centering
\includegraphics[width=1.0\textwidth]{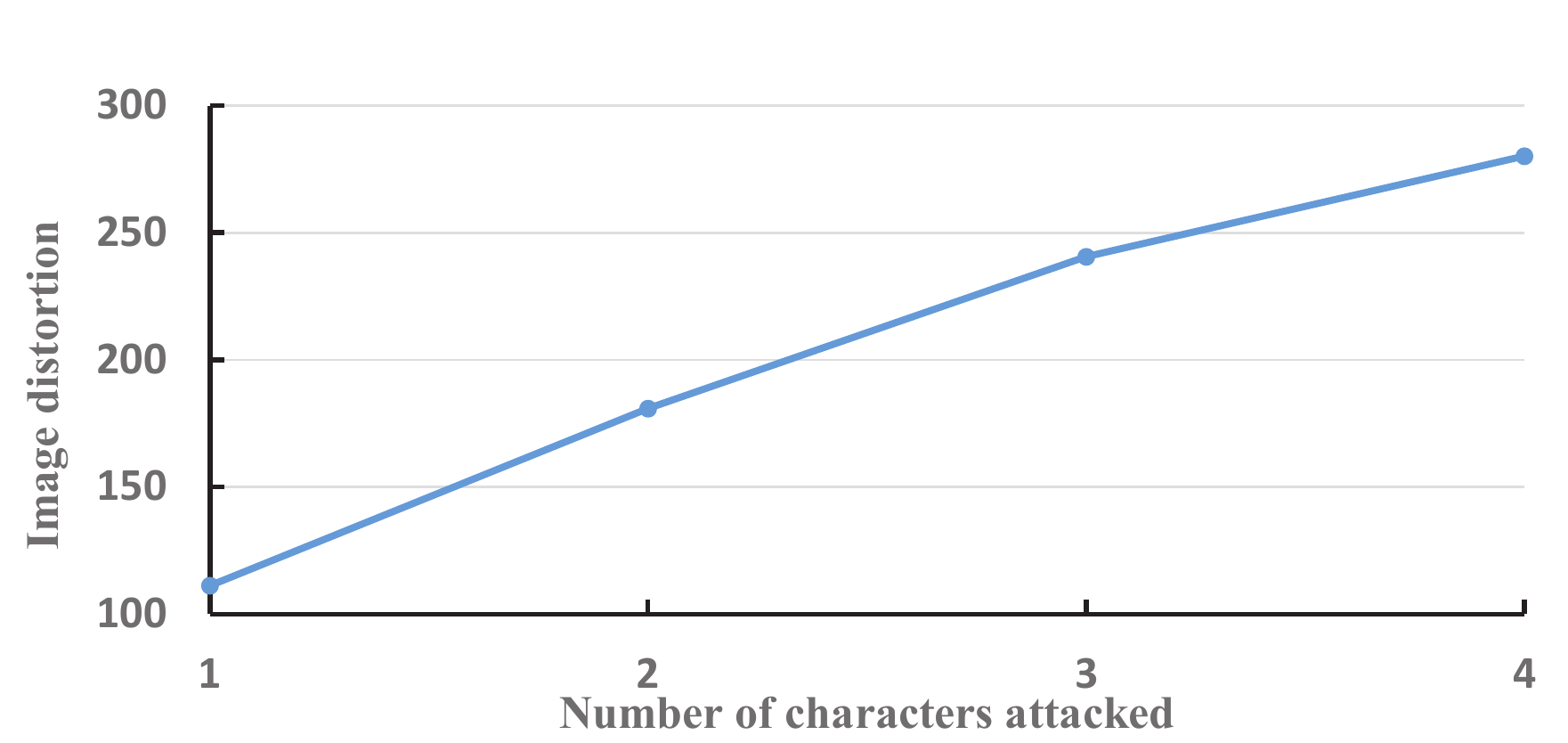}
\centerline{(a) Image distortion}
\end{minipage}
\begin{minipage}[b]{0.99\linewidth}
\centering
\includegraphics[width=1.0\textwidth]{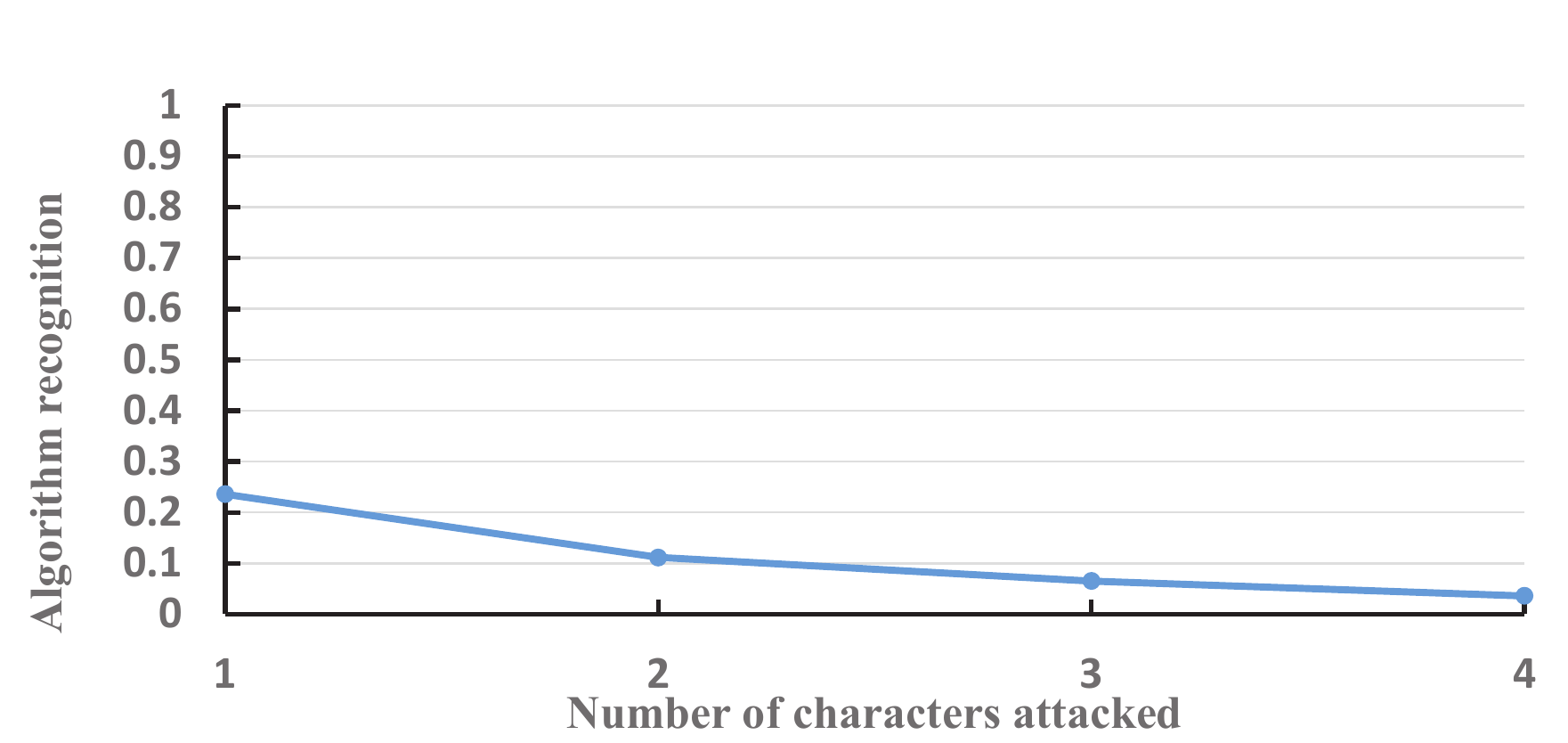}
\centerline{(b) Algorithm recognition accuracy}
\end{minipage}
\caption{The influence of $|\Theta|$ on derived image distortion and cracking recognition accuracy.}
\label{fig:tokens}
\end{figure}

\begin{table*}[t]
  \centering
  \caption{Transferability of adversarial images generated between pairs of models. The element ($i, j$) represents the accuracy of the adversarial images generated for model $i$ (row) tested over model $j$ (column).}\label{tab:3}
  \begin{tabular}{|c|c|c|c|c|} 
  \hline 
   &$4$ConvNet& ResNet& DenseNet& GoogLeNet\\
  \hline  
  $4$ConvNet& 1\%& 3\%& 13\%& 7\%\\
  \hline 
  ResNet& 8\%& 0\%& 13\%& 12\%\\
  \hline
  DenseNet& 16\%& 2\%& 3\%& 23\%\\
  \hline
  Ensemble Training& 0\%& 2\%& 1\%& 3\%\\
  \hline
  \end{tabular}
\end{table*}

\subsection{Parameter Influence Analysis}\label{sec4-3}

The proposed robust CAPTCHA generation method mainly involves with two parameters: the weight parameter $\lambda$ in Eqn.~\ref{eq1} and the number of attacked characters $|\Theta|$.
As introduced in the methodology, the weight parameter $\lambda$ controls the relative importance between the visual distortion and misclassification confidence. We adjusted $\lambda$ within range of $[10,30]$ with step of $1$ and examined its influence on the derived image distortion and cracking recognition accuracy. The image distortion is measured as the sum of squared pixel-wise difference between the original and adversarial images. The averaged distortion and recognition accuracy with the change of $\lambda$ are drawn in Fig.~\ref{fig:lambda}. It is shown that as $\lambda$ increases, more image distortion is observed in the derived CAPTCHA images and cracking method tends to fail in recognizing the generated CAPTCHAs. This is consistent with the definition of $\lambda$ in Eqn.~\ref{eq1}. In practical applications, to prevent annoying human subjects in recognizing the generated CAPTCHAs, an appropriate $\lambda$ is selected with moderate image distortion and guaranteed cracking resistent performance. Our experimental results reported in Section~\ref{sec4-2} are based on $\lambda=20$.

Regarding the number of attacked characters, we set $|\Theta|$ to $\{1,2,3,4\}$ respectively and examined the corresponding averaged image distortion and algorithm recognition accuracy in Fig~\ref{fig:tokens}. As shown in Fig~\ref{fig:tokens}(a), with $|\Theta|$ increases, more image distortion is needed to misclassify the characters. In Fig.~\ref{fig:distortion} we show example CAPTCHAs generated by rCAPTCHA with different levels of image distortions. Combing with Fig~\ref{fig:tokens}(a), it is demonstrated that using the proposed rCAPTCHA method to attack even all $4$ characters, the derived CAPTCHAs are generally friendly to human and not bringing extra recognition burden. As shown in Fig~\ref{fig:tokens}(b), the increase of $|\Theta|$ enhances the confidence to mislead the cracking algorithm and obtains consistently lower recognition accuracy. With the introduction of multi-attack towards sequential recognition, the proposed rCAPTCHA method possess the flexibility to attack arbitrary number of characters. In our experiments, to guarantee the resistent capability, we fixed $|\Theta|=4$.

\subsection{Robustness towards Different OCRs}\label{sec4-4}

To justify the generalization and transferability of our proposed rCAPTCHA method, we implemented different cracking methods and examined their recognition accuracy on the generated CAPTCHAs. Specifically, we respectively trained $4$ OCR models with different CNN structures, which are denoted as \emph{$4$ConvNet}, \emph{mini-ResNet}, \emph{mini-DenseNet} and \emph{mini-GoogLeNet}. \emph{$4$ConvNet} uses four convolutional layers for feature extraction. \emph{mini-XNet}s are employed due to the low resolution of CAPTCHA images: \emph{mini-ResNet} consists of five ResBlocks and two convolutional layers, \emph{mini-DenseNet} consists of four DenseBlocks with four convolutional layers, and \emph{mini-GoogLeNet} consists of two Inception modules with six convolutional layers.

Among the $4$ models, $3$ models of  \emph{$4$ConvNet}, \emph{mini-ResNet}, \emph{mini-DenseNet} are selected as white-boxs, with the remaining \emph{mini-GoogLeNet} model as the black-box. The black-box model is regard as the potential OCR cracking to simulate the alternative cracking choices in real-world applications. Averaged over $100$ tested CAPTCHAs, Table~\ref{tab:3} summarizes the black-box cracking recognition accuracy under different training-testing pairs. For example, the value of $0\%$ at element $(1,4)$ indicates the recognition accuracy trained with ensembled $3$ white-box models and tested on \emph{$4$ConvNet}. Lower accuracy value means superior resistent performance to cracking solutions and better transferability of the method. We observe that the adversarial images generated with one model perform well on their own models (diagonal elements) but generally perform poorly on other models. However, if we generate the CAPTCHA images with ensemble training of $3$ models, the testing recognition accuracies for all $4$ models are no higher than $3\%$ (the last row). This demonstrates the transferability of the proposed rCAPTCHA method in employing ensemble training towards black-box cracking. It is expected with more models implemented in ensemble training, the resistent performance towards arbitrary black-box cracking methods will be guaranteed. In practical applications, we can carefully select white-box models with typically different structures to improve the generalization and transferability to specific models.

\section{CONCLUSION}

This study designs robust character-based CAPTCHAs to resist cracking algorithms by employing their unrobustness to adversarial perturbation. We have conducted data analysis and observed human and algorithm's different vulnerabilities to visual distortions. Based on the observation, robust CAPTCHA (rCAPTCHA) generation framework is introduced with three modules of multi-target attack, ensemble adversarial training, and differentiable approximation to image preprocessing. Qualitative and quantitative experimental results demonstrate the effectiveness of generated CAPTCHAs in resisting cracking algorithms.

It is noted that similar to the game competition between adversarial attack and defense, with more CAPTCHA designers employing adversarial attack to resist cracking, future cracking solutions are expected to employ adversarial defense techniques for self-enhancement. We hope this study could draw attention of future CAPTCHA designing on the competition between adversarial attack and defense. Moreover, with the development of deep learning and other AI algorithms, we are confronted with critical security-related problems when algorithms are maliciously utilized towards human. In this case, it is necessary to get aware of the limitations of current algorithms and appropriately employ them to resist the abuse use of algorithms.




{\small
\bibliographystyle{ieee}
\bibliography{sample-base}
}

\end{document}